\documentclass[preprint,12pt]{elsarticle}

\usepackage{graphicx, subfigure} 
\usepackage{booktabs}
\usepackage{graphicx}%
\usepackage{ulem}
\usepackage{multirow}%
\usepackage{amsmath,amssymb,amsfonts}%
\usepackage{amsthm}%
\usepackage{mathrsfs}%
\usepackage{subcaption}%
\usepackage{makecell}%
\usepackage{url}
\usepackage{comment}
\usepackage{array}
\usepackage{lineno}%
\usepackage{xcolor}%
\usepackage{pdflscape} 

\journal{Ecological Informatics}

\begin{document}

\begin{frontmatter}
\title{Low Cost, High Efficiency: LiDAR Place Recognition in Vineyards with Matryoshka Representation Learning
} 
\author[elche]{Judith Vilella-Cantos\corref{cor1}} 
\ead{jvilella@umh.es}
\author[torino]{Mauro Martini}
\author[torino]{Marcello Chiaberge}
\author[elche]{Mónica Ballesta}
\author[elche]{David Valiente}
\affiliation[elche]{organization={University Institute for Engineering Research, Miguel Hernández University},
        addressline={Avda. de la Universidad s/n, Edificio Innova}, 
            city={Elche},
            postcode={03202}, 
            state={Alicante},
            country={Spain}}
\affiliation[torino]{organization={Department of Electronics and Communications, Politecnico di Torino},
            city={Torino},
            postcode={10129}, 
            country={Italy}}
\cortext[cor1]{Corresponding author}
\begin{abstract}
    Localization in agricultural environments is challenging due to their unstructured nature and lack of distinctive landmarks. Although agricultural settings have been studied in the context of object classification and segmentation, the place recognition task for mobile robots is not trivial in the current state of the art. In this study, we propose MinkUNeXt-VINE, a lightweight, deep-learning-based method that surpasses state-of-the-art methods in vineyard environments thanks to its pre-processing and Matryoshka Representation Learning multi-loss approach. Our method prioritizes enhanced performance with low-cost, sparse LiDAR inputs and lower-dimensionality outputs to ensure high efficiency in real-time scenarios. Additionally, we present a comprehensive ablation study of the results on various evaluation cases and two extensive long-term vineyard datasets employing different LiDAR sensors. The results demonstrate the efficiency of the trade-off output produced by this approach, as well as its robust performance on low-cost and low-resolution input data. The code is publicly available for reproduction on a Github repository.\footnote{\url{https://github.com/JudithV/MinkUNeXt-VINE}}
\end{abstract}

\end{frontmatter}

\section{Introduction}
In the domain of robotic localization, the place recognition problem involves the ability to accurately match a place on real-time navigation in a database which stores the map of previously visited locations. A robust solution to this problem allows any robotic system to navigate the environment accurately and successfully localize itself in GPS-denied environments. This is crucial in every simultaneous localization and mapping (SLAM) application and in loop closure detection algorithms \cite{kazerouni2022survey, macario2022comprehensive, zhang20243d}.

The utilization of Light Detection and Ranging (LiDAR) technology as an input in these algorithms is a growing approach. The 3D data from scans are more robust to illumination and weather changes than traditional vision systems. Therefore, applying deep learning techniques to point clouds for place recognition tasks has become a critical component in modern autonomous navigation systems \cite{uy2018pointnetvlad, komorowski2021minkloc3d, chen2022overlapnet}. 

Agricultural environments present a series of unique challenges that are not present in urban settings, which are the most widely tested environments. In this study, the research is conducted in a specific agricultural context: vineyards. Among the aforementioned challenges, the presence of repetitive structures without distinctive landmarks, occlusions by heavy vegetation \cite{lehnert20193d}, and significant changes in the appearance of the scene between different seasons are particularly prominent \cite{ding2022recent}. Furthermore, the geometry of vine rows produces limited discriminative features for both visual and LiDAR sensors.

The identification of a robust solution to the LiDAR place recognition (LPR) problem in agricultural settings has the potential to facilitate the development of advanced applications in precision agriculture \cite{coulibaly2022deep}, field and crop monitoring \cite{teng2023multimodal}, phenotyping \cite{chebrolu2021registration}, and autonomous navigation of agricultural vehicles \cite{padhiary2024enhancing}. In order to address the specific challenges posed by vineyard environments, a series of experiments were conducted with different backbones and scopes of the training and evaluation data. We analyze how repetitive plant structures and seasonal variability affect the robustness of current place recognition networks, and propose strategies to improve generalization in such challenging agricultural settings. Finally, we propose MinkUNeXt-VINE, a variant of the MinkUNeXt backbone \cite{cabrera2025minkunext}, known for its strong performance in urban LPR. We redesign the original architecture by pruning inefficient layers and enforcing a lower-dimensional output, which drastically lowers parameter count and latency. We couple this lightweight design with Matryoshka Representation Learning (MRL) \cite{kusupati2022matryoshka}, a training strategy that enforces structure at various embedding sizes via a weighted summation of descriptor slices. We evaluate the performance of our approach (defined herein as retrieval accuracy via Recall metrics) and its efficiency (computational overhead and descriptor size) across several datasets. The resulting framework offers a scalable and efficient solution designed for real-time deployment in agricultural localization.

The contributions of our work are the following:
\begin{itemize}
    \item We introduce MinkUNeXt-VINE, a lightweight LPR network designed for agricultural robotics. It leverages a multi-loss strategy through MRL and architectural simplification to produce flexible, low-dimensional descriptors suitable for real-time constraints. This is the first contribution using a MRL approach for optimizing a LPR method.
    \item We conduct a thorough analysis of architectural trade-offs, specifically investigating the synergy between MRL, reduced layer complexity, and descriptor dimensionality, proving the performance of our pruning strategy in inputs such as a Livox LiDAR sensor (low-cost, low-resolution).
    \item We present an extensive validation of MinkUNeXt-VINE in cross-season scenarios, testing the robustness of our method across the different months and seasons conforming the phenological cycle of the vineyards.
    \item A complete benchmark of LiDAR place recognition approaches over most recent long-term vineyard datasets. We establish a comparison between the impact of our method on different cost and resolution LiDAR sensors such as Livox and Velodyne.
\end{itemize}

\section{Related works}
\subsection{Place recognition in agricultural environments}
Agricultural environments are distinguished by their uniformity and high variability in the different seasons of the year. The environment undergoes notable changes depending on the crop's developmental stage, including periods of high vegetation density within the field, which can impede the functionality of robotic vision systems. In particular, within vineyard environments characterized by dense foliage, the robot's field of vision is constrained to the immediate row in which it is operating. 

The issue of place recognition in these environments remains an active area of research. While classification and semantic segmentation have been the focus of these environments, the realm of localization remains largely unexplored. With regard to the issue of vision localization, proposed solutions include an approach by Liu et al. \cite{liu2023vision} for autonomous navigation in vineyard environments. This approach utilizes a path detection framework that detects traversability from the RGB-D image source. In \cite{aghi2020local}, the authors propose a localization solution that uses an estimation of the location of the end of the vineyard row through a depth map and a modified MobileNet \cite{howard2017mobilenets} for the classification of objects detected in the vision system. This approach facilitates the localization process and was tested in real-world environments.

In the case of the LPR problem, several recent works have addressed it on agricultural environments. In TriLoc-NetVLAD \cite{sun2024triloc}, the authors propose an orchard-focused, long-term LPR solution consisting of a spatial context descriptor based on point cloud density, height, and spatial information. Then, they use a triplet neural network that fuses high-dimensional and low-dimensional features with a NetVLAD \cite{arandjelovic2016netvlad} aggregation module. The methodology in SPVSoAP3D \cite{barros2024spvsoap3d} consists of a voxel-based feature extraction module (SPV), followed by a feature extraction module that combines second-order pooling, log-Euclidean projection, and power normalization (SoAP). Finally, it produces the descriptor through a fully connected layer with L2 normalization. In a similar vein, PointNetPGAP-SLC \cite{barros2024pointnetpgap} proposed by Barros et al. presents a methodology for the extraction of features from point clouds by leveraging the capabilities of PointNet \cite{qi2017pointnet}. The subsequent aggregation of these features is achieved through a process that integrates the Pairwise Feature Interactions and Global Average Pooling (PGAP) aggregation techniques. To enhance the robustness of this methodology, a signal that considers the segmentation of vineyards is incorporated into the Segment-Level Consistency (SLC) model.

\subsection{LiDAR place recognition}
The use of LiDAR technology instead of a traditional 2D vision system provides invariance to lightning changes. For this reason, the development of algorithms that take as input these 3D data has been increasing over the last years. The grounding work for the development of LPR deep learning algorithms is PointNetVLAD \cite{uy2018pointnetvlad}. This method combines the use of the PointNet architecture \cite{qi2017pointnet} to extract higher dimensional local features from the point clouds and pass them to a NetVLAD \cite{arandjelovic2016netvlad} aggregation module. After the creation of PointNetVLAD, several works approached the LPR problem with deep learning techniques. In LPD-Net \cite{liu2019lpd}, the network incorporates additional contextual information regarding the neighborhood of each point. This additional information encompasses the linearity, planarity, and scattering properties of the neighborhood. In contrast to other networks, LPD-Net utilizes this expanded set of features for feature extraction, in addition to positional information. Instead of NetVLAD, a graph-based aggregation module is utilized. In MinkLoc3D \cite{komorowski2021minkloc3d}, the authors propose a method based on Minkowski convolutions \cite{choy20194d}, which operates on sparse tensors of unordered 3D coordinates. On the same note, MinkUNeXt \cite{cabrera2025minkunext} is a deep network that achieves exceptional recall results on the LPR tasks, using only Minkowski convolutions and transposed Minkowski convolutions without additional complex mechanisms such as transformers or attention modules.

Apart from the deep learning approaches to the LPR problem, some methods use handcrafted global descriptors to solve it. These descriptors are not trained through a machine learning system; rather, they are built from the geometric characteristics of LiDAR scans. Scan Context \cite{kim2018scan} is a popular example. It creates a global descriptor by dividing the original point cloud into sectors and voxels. Then, it encodes the descriptor by taking into account the maximum height of the points in each bin. Other methods such as Fusion Scan Context \cite{deng2023fusion} also take into account the density of each bin (how many points are present in each of them) and the intensity of each bin. In Semantic Scan Context \cite{li2021ssc} is augmented by the application of semantic segmentation to the raw point cloud, with the resultant information projected onto the plane. A two-step global semantic ICP is also performed to obtain the 3D pose and to robustly align the point clouds. A distinct handcrafted place recognition descriptor is the Stable Triangle Descriptor (STD) \cite{yuan2023std}. The STD is a six-dimensional vector formed by the length of the three triangle sides and the three normal vectors corresponding to each of the three vertices of the triangle. It is imperative to note that the term "triangle" in this context refers to a geometric figure delineated by three keypoints, which are identified within the point cloud subsequent to its projection onto a plane. The recent Omni Point method \cite{im2024omni}, as outlined in the study, builds a descriptor from the distance between the feature points present in the point cloud, following a histogram of distances approach.

LiDAR-based place recognition is a promising field due to the immutability of these geometric data to external influences. However, further research is necessary to extend these solutions to agricultural environments and obtain successful results. For instance, the SPVSoAP3D \cite{barros2024spvsoap3d} method obtains a Recall@1 of 76.30\% in the most favorable agricultural case, while an urban method such as MinkUNeXt \cite{cabrera2025minkunext} achieves 95.80\% for the same metric across all tested protocols. This sharp contrast clearly highlights the need for further research to successfully adapt these solutions to agricultural environments.

\section{Methodology}\label{sec:methodology}
To address the unique challenges of vineyard environments and low-cost LiDAR sensors, we propose MinkUNeXt-VINE. It is a lightweight LPR pipeline optimized for repetitive settings with a focus on high efficiency, based on the backbone MinkUNeXt \cite{cabrera2025minkunext}. 

To provide an exhaustive explanation of the proposed method, Section \ref{subsec:input_prepro} presents, firstly, the preprocessing steps taken to stabilize the sparse input for our method. Section \ref{subsec:minkunext-vine} discusses the architectural decisions that led us to our final proposal, MinkUNeXt-VINE. Finally, Section \ref{subsec:MRL} provides details about the MRL multi-loss strategy adopted by our algorithm.



\subsection{Input preprocessing}\label{subsec:input_prepro}
In order to ensure a stable, reliable and higher performance, we applied preprocessing steps to the raw points extracted from the point clouds.

In order to ensure the integrity of our algorithm, a process of filtration is implemented to remove points that may contain extraneous data, such as those characterized by zero values for the x, y, z coordinates. Additionally, a filtration process is employed to remove points that are situated at a distance greater than $60$ meters, a procedure that is applied to both datasets. Finally, we normalize each point in a $[-1, 1]$ range. This proposed global normalization process involves two stages: fixed-range scaling and zero-centering. First, the coordinates are scaled by a fixed factor $S=60$, mapping the mapped range of the sensor to the normalized interval $[-1, 1]$. This transformation preserves the geometric aspect ratio while standardizing the input variance. Subsequently, the point cloud is translated to the origin by subtracting the centroid (mean) from each coordinate, effectively removing translation invariance. Equation \ref{eq:global_norm} illustrates this global normalization process.

\begin{equation}
    \hat{p}_i = \frac{p_i - \mu}{S}
    \label{eq:global_norm}
\end{equation}
Where $\hat{p}_i$ represent the normalized original $p_i$ point, $\mu$ the centroid of the point cloud and $S$ the normalization factor which, as previously mentioned, is equivalent to $60$. 

\subsection{MinkUNeXt-VINE}\label{subsec:minkunext-vine}
MinkUNeXt \cite{cabrera2025minkunext} is a LiDAR place recognition method that follows a U-Net architecture and relies solely on Minkowski convolutions to produce a robust global descriptor. This network is based on the encoder-decoder type, incorporating skip connections. The utilization of skip connections facilitates the integration of low-level and high-level features, thereby enhancing the performance of the place recognition task. 


Building upon the insights of Máximo et al. \cite{maximo2025coarse} that descriptors emerging after the second transpose convolution yield minimized mean positional and orientation errors, we designated this layer's output as the final feature map. We then pruned the subsequent five layers and the final fully connected layer from the original backbone. This resulted in a 192-dimensional global descriptor, which is much more compact than the original 512-dimensional global descriptor. As demonstrated in Section \ref{subsec:multiloss_design}, this streamlining significantly reduces pipeline complexity while enhancing retrieval performance.

\begin{figure}[h]
    \centering
    \includegraphics[width=\linewidth]{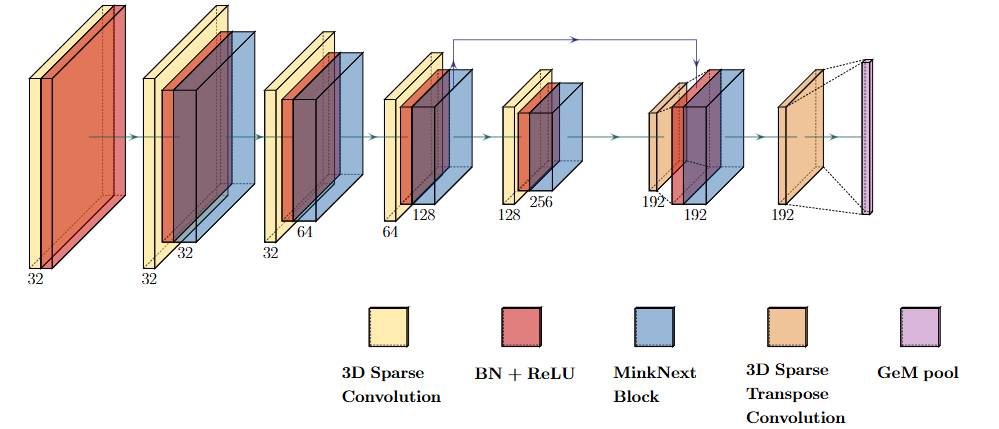}
    \caption{MinkUNeXt-VINE's proposed architecture. The parent backbone has been pruned to improve efficiency and performance in unstructured settings.}
    \label{fig:full_pipeline}
\end{figure}

Figure \ref{fig:full_pipeline} depicts the architecture of MinkUNeXt-VINE, highlighting the strategic integration of the GeM pooling aggregator immediately following the second transpose convolution. This design yields a significantly more compact pipeline compared to the baseline, ensuring efficient real-time performance as quantitatively demonstrated in Section \ref{subsec:mem_analysis}.

\subsection{Learning with a multi-loss approach}\label{subsec:MRL}
The original MinkUNeXt method uses a simple Truncated Smooth Average Precision (TSAP) loss function, which aims to maximize the ranking of the positives with the following formula:
\begin{equation}
    \mathcal{L}_{\text{TSAP}}=\frac{1}{b}\sum_{q=1}^{b}(1-AP_q)
\end{equation}
Where $b$ is the batch size and $AP$ is the smooth average precision, which is calculated with this formula:
\begin{equation}
    AP=\frac{1}{|P|}\sum_{i\in P}\frac{1+\sum_{j\in P, j\neq i}G(d(q,i)-d(q,j); \tau)}{1+\sum_{j\in \Omega, j\neq i}G(d(q,i)-d(q,j); \tau)}
\end{equation}
Where, for a set of positives $P$, the average precision of a given point cloud $q$ is computed as the mean of the soft precision values calculated for each positive sample $i$. Here, $\Omega$ represents the complete set of all samples in the batch, including both positives and negatives. On the other hand, $d(\cdot, \cdot)$ is the distance metric between the query descriptor and a candidate, represented by the Euclidean distance. The Sigmoid function $G(\cdot; \tau)$, where $\tau$ is a temperature parameter, controls the sharpness of the ranking approximation. 

Nevertheless, multi-loss approaches have been proven to enhance feature robustness by enforcing constraints at different levels of granularity \cite{liu2024novel, wazir2022histoseg, thi2025multi}. A specific multi-loss paradigm that has gained significant traction recently is Matryoshka Representation Learning (MRL) \cite{kusupati2022matryoshka}. Mimicking the structure of nesting dolls, MRL enforces that nested sub-vectors of the full descriptor independently encode discriminative information. This is achieved through a weighted summation strategy, where the loss is evaluated at multiple embedding dimensions simultaneously.

The formula for the MRL approach adopted by our method is defined in Equation \ref{eq:mrl}:

\begin{equation} \mathcal{L}_{\text{mat}} = \sum_{m \in \mathcal{M}} w_m \cdot \mathcal{L}_{\text{TSAP}}(\mathbf{z}_{1:m}) \label{eq:mrl} \end{equation}

where $\mathcal{L}_{\text{mat}}$ represents the total combined loss. The term $\mathcal{M}$ denotes the set of nested dimensions (e.g., $\{16, 32, 64, ..., d\}$), $w_m$ is the weight assigned to each granularity, and $\mathcal{L}_{\text{TSAP}}(\mathbf{z}_{1:m})$ is the TSAP loss computed on the first $m$ dimensions of the descriptor $\mathbf{z}$. In our MinkUNeXt-VINE, we select the list of dimensions $[64, 128, 192]$ with the weights of $[1.0, 0.5, 0.25]$ for each dimension, based on the results of the preliminary dimensionality ablation study presented in Section \ref{subsec:dimensionality_results}. This approach prioritizes lower-dimensionalities to focus on high efficiency.

\section{Datasets}\label{sec:datasets}
Traditionally, public data from vineyard environments have been centered around vision tasks (classification and segmentation tasks) \cite{barros2022multispectral}. For example, the GrapesNet dataset \cite{barbole2023grapesnet} provides RGB-D imagery taken on a vineyard located in India. Abdelghafour et al. \cite{abdelghafour2021annotated} also presented an annotated imagery dataset for vineyard environments. The aforementioned datasets were published with the objective of addressing the necessity for data classification in the context of disease identification tasks.

In this study, we use the data from the only two vineyard datasets to this date that are taken in long-term periods with 3D LiDAR sensors: the Bacchus Long-Term (BLT) Dataset \cite{polvara2024bacchus} and the TEMPO-VINE dataset \cite{martini2025tempo}.

\subsection{Bacchus Long-Term Dataset}\label{subsec:dataset_blt}
The BLT dataset \cite{polvara2024bacchus} was collected in two different vineyard environments located in Greece (Ktima Gerovassiliou) and the United Kingdom (Riseholme). The Greek campaigns are eleven, recorded between the months of March and September and covering all the different phases of the crop. On the other side, the campaigns taken in the Riseholme vineyard are five and were taken in the months of July and August. It is important to note that, in the initial two campaigns conducted at the Riseholme vineyard, GPS data is not available. Consequently, these campaigns are not appropriate for LPR algorithms due to the absence of a ground-truth source.

With regard to the expansion of the data set collected in this particular vineyard, the robot traverses a total of five rows within the Ktima Gerovassiliou vineyard, with each row spanning a length of 50 meters. Similarly, the Riseholme vineyard encompasses five rows, which are half the length of their Greek counterpart. Both data collection settings encompass the process of revisiting the same row in reverse order.

Regarding the robotic setup employed to collect these data, a robotic SAGA Robotics Thorvald II platform was used. The sensor mounting includes two Zed2i stereo RGB-D cameras, a SICK MRS1000 2D LiDAR, a FPS BX992 GPS-RTK system, a RSK-UM7 inertial measurement unit (IMU) sensor and an Ouster OS1-16 3D LiDAR. For the LPR task, the Ouster OS1-16 provides high quality point clouds of the environment, as it is able to capture more than 300k points per second with a range of more than 100 meters with its 16 simultaneous beams.

\begin{figure}[h]
    \centering
    \includegraphics[width=\textwidth]{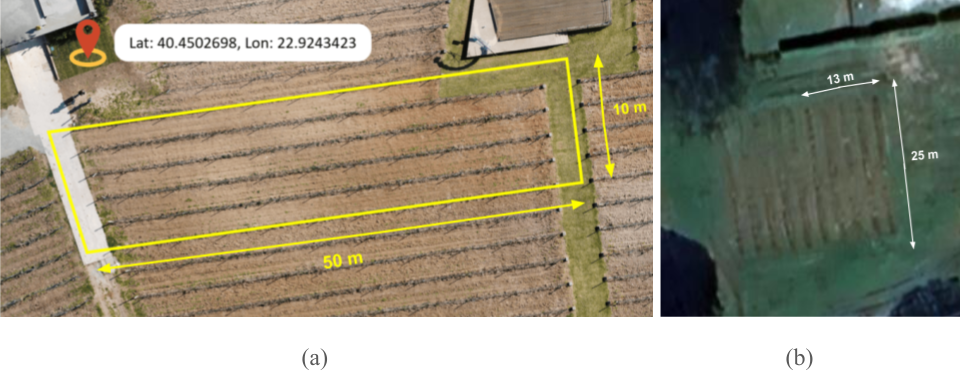}
    \caption{Extension of each of the vineyards considered within the BLT datasets \cite{polvara2024bacchus}. a) Ktima Gerovassiliou (Greece) b) Riseholme (United Kingdom)}
    \label{fig:blt_dataset}
\end{figure}

Figure \ref{fig:blt_dataset} depicts the two vineyards that comprise the BLT dataset with their corresponding extension.

\subsection{TEMPO-VINE dataset}\label{subsec:dataset_tv}
The TEMPO-VINE dataset \cite{martini2025tempo} was collected in northern Italy (Agliè) in two distinct vineyard settings: pergola and trellis. The campaigns conducted at this dataset are a total of eleven, starting in February and finishing in October, which results in a total coverage of all seasons and crop stages. For each campaign, two runs are collected in the pergola setting and two runs are collected in the trellis setting. Although the two runs recorded in the pergola cover the same path, there is a discrepancy between the two runs captured in the trellis environment. Specifically, "run1" encompasses 10 rows, while "run2" encompasses 8 rows. The first four rows of "run2" align with the first four rows of "run1," while the subsequent four rows traverse in the opposite direction. Furthermore, from the latter campaigns conducted in June onward, a third row was collected in the trellis environment, encompassing revisits to the first three rows of the trellis in reverse. Taking this into account, the total number of trajectories available of this dataset is twenty-four. 

Each row in both the pergola and trellis settings is 110 meters long. This dataset's distinctive features are its captions of the diverse growth stages of grass in a field and the fact that the dataset was recorded at a linear speed of one meter per second. These attributes introduce an additional layer of complexity to the place recognition problem, making it more difficult to identify the exact location within the field.

\begin{figure}[h]
    \centering
    \includegraphics[width=\linewidth]{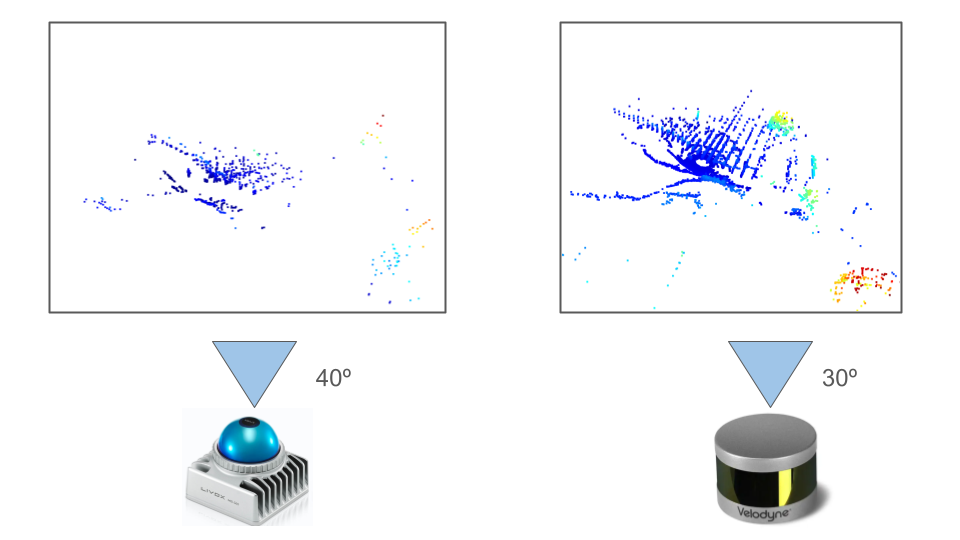}
    \caption{Comparison between the two LiDAR sensors used in the TEMPO-VINE dataset collection. Left: Point cloud captured with the Livox. Right: Point cloud captured with the Velodyne. The vertical FoV is annotated next to each sensor, while the horizontal FoV is 360$^{\circ}$ for both.}
    \label{fig:lidar_comparison}
\end{figure}

Regarding the robotic setup of this dataset, a Husky UGV platform from Clearpath Robotics was used. The sensors collected include a RGB-D Intel Realsense D435 camera, a Swift Navigation Duro GPS-RTK system and a MicroStrain 3DM-GX5 IMU. The dataset was collected using two different LiDARs: a Velodyne VLP-16 and a Livox MID360. The two sources under consideration yield point clouds of varying characteristics. The Livox employs a non-repetitive scanning pattern, whereby the point clouds are constructed from the exposure time. In contrast, the Velodyne adheres to a conventional time-of-flight (TOF) pattern. Consequently, the Livox sensors generate considerably less detailed point clouds, a common feature in lower cost sensors. Figure \ref{fig:lidar_comparison} illustrates the point clouds generated by both sensors captured simultaneously, highlighting the FoV characteristics of each device. Notably, the point cloud obtained with the Velodyne sensor contains approximately 10,000 more points per frame than its Livox counterpart.

\begin{figure}[h]
    \centering
    \includegraphics[width=0.75\linewidth]{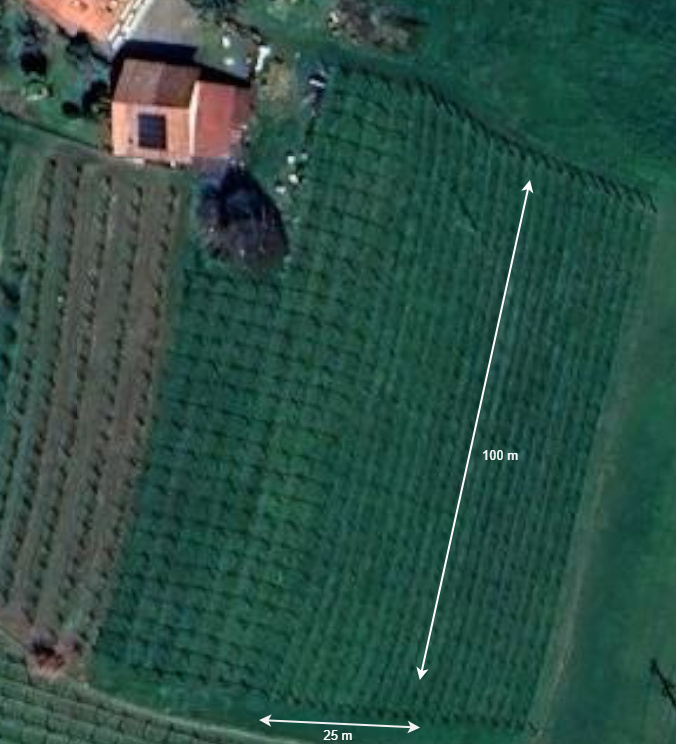}
    \caption{Extension of the field where the TEMPO-VINE dataset \cite{martini2025tempo} was recorded.}
    \label{fig:tempo-vine-dataset}
\end{figure}

Figure \ref{fig:tempo-vine-dataset} displays the field corresponding to the TEMPO-VINE dataset. The pergola rows are situated in the left portion of the field, while the rows corresponding to the trellis are located in the central and right parts of the image. It is important to note that, due to the curvature of the field, the extension values are approximate (the rows on the left side are longer than those on the right side). In the pergola scenario, the rows are wider (3.5 meters), while in the trellis, the rows have a width of 2 meters. The pergola section measures 15 meters in width, with a length of 110 meters that aligns with the westernmost section of the trellis.

\section{Experimental setup}\label{sec:experimental_setup}
In order to assess the performance of the proposed MinkUNeXt-VINE, a series of experiments were conducted on two distinct datasets in terms of recall, as outlined in Section \ref{sec:datasets}. The results in both Section \ref{sec:ablation_study} and \ref{sec:results} are expressed via the Recall@1 and Recall@1\% metrics. Throughout our manuscript, we refer as performance to the accuracy of the place recognition algorithm through the Recall@1 and Recall@1\% metrics. The recall value measures the performance of a place recognition algorithm as follows:

\begin{equation}
    Recall = \frac{TP}{TP + FN}
\end{equation}

Being $TP$ the amount of the true positives and $FN$ the amount of false negatives found during the evaluation phase of the algorithm.

In Section \ref{sec:results}, the division of the query into a training set and a testing set is executed in accordance with the methodology outlined by Uy et al. in their PointNetVLAD contribution \cite{uy2018pointnetvlad}. The trajectory of the vehicle is subdivided into a series of zones, defined by a predetermined set of points. These points are determined by the robot's GPS position and a radius in meters, which is used to delineate the area of interest. A subset of zones within this trajectory is reserved for testing only, while the remainder is incorporated into the training set. For the BLT, the radius was set to 13 meters, and three reference points for the test zones were chosen in different areas of the vineyard. For the TEMPO-VINE dataset, the radius was set to 23 meters, and four reference points for the test zones were selected in both the pergola and trellis sections of the vineyard.

For the results displayed on Section \ref{sec:ablation_study}, unlike the zone-based division method used for Section \ref{sec:results}, we conducted experiments using two additional types of sampling. The results in Section \ref{sec:ablation_study} were obtained using only the Livox data from the TEMPO-VINE dataset, as our research focuses primarily on low-cost sensors. The first approach involves implementing a methodology that aligns with the dataset's structural characteristics. In this methodology, the "run2" trajectory is designated as the test set. Construction of the test set comprises 40 to 50 percent of the examples and is contingent upon inclusion of the "run3" trajectory within the campaign. The second approach uses a "one in, one out" division. In this approach, one query is first entered into the training set, then the next query is entered into the test set, and so on. Thus, the training set comprises 50\% of the total set of examples, and the test set consists of the remaining 50\%. 

All three sensors that were recorded between these two datasets have a sampling frequency of 10 Hz. To avoid redundant information and enhance computational efficiency, sampling is based on distance. For the BLT dataset, since the area is considerably smaller, scans collected every half meter are considered. Conversely, for the TEMPO-VINE dataset, scans are considered every meter.

\section{Ablation study}\label{sec:ablation_study}
This section details the ablation studies conducted to substantiate the architectural choices of MinkUNeXt-VINE for efficient vineyard place recognition. We begin by analyzing the trade-off between performance (Recall metrics) and efficiency (computational latency and storage requirements). Specifically, we vary the output dimensionality of the global descriptor; while lower-dimensional descriptors enhance efficiency by reducing memory footprint and search time, they may sacrifice the discriminative power necessary for high performance. Subsequently, we evaluate the impact of the loss function, specifically highlighting the benefits of adopting a MRL approach. Apart from evaluating different approaches of loss function selection and adjustments, we also ponder the impact of choosing the final layer of the network from different points: MinkUNeXt's final layer or our proposed pruned architecture. Finally, to address the challenge of long-term autonomy, we assess the model's generalization capabilities across different months and quarterly data partitions. All experiments reported in this section were conducted on an Nvidia A30 GPU. 

\subsection{Dimensionality study}\label{subsec:dimensionality_results}
While urban settings are commonly rich in unique semantic landmarks, agricultural scenes suffer from feature scarcity and self-similarity. Consequently, generating a global descriptor in this domain presents a trade-off: broadly descriptive vectors may capture unnecessary high-frequency noise, while overly compressed vectors might lose discriminative power. This ablation study aims to determine the optimal dimensionality that balances feature robustness against noise suppression, ensuring an accurate depiction of the environment.

Table \ref{tab:dimensionality_results} displays the results for the MinkUNeXt method with different output dimensions on the Livox point clouds from the TEMPO-VINE dataset. The setup for these experiments consists of the original MinkUNeXt architecture with a quantization size value of $0.01$ meters and the normalization of the points in a $[-1, 1]$ range. The global normalization applied to the points was made following the procedure exposed under Section \ref{subsec:input_prepro}. The evaluation of these results was conducted in three distinct cases: firstly, an assessment was made against the same month (February); secondly, an evaluation was conducted against May; and thirdly, an evaluation was conducted against the second campaign of June. The two latter cases demonstrate the impact of the descriptor's dimensionality on cross-season evaluations. 

\begin{table}[h]
    \centering
    \scriptsize
    {
    \begin{tabular}{|c|c|c|c|c|c|c|}
        \hline
        \multirow{2}{*}{\textbf{\makecell{Output\\dimensions}}} & \multicolumn{2}{c|}{\textbf{February}} & \multicolumn{2}{c|}{\textbf{May}} & \multicolumn{2}{c|}{\textbf{June}} \\
        \cline{2-7}
        & Recall@1\% & Recall@1 & Recall@1\% & Recall@1 & Recall@1\% & Recall@1 \\
         \hline
         512 & 95.59 & 58.82 & 67.15 & 21.81 & 70.34 & 25.61\\
         256 & 95.59 & 58.82 & \textbf{68.88} & 20.61 & \textbf{71.94} & 25.00\\
         192 & \textbf{95.83} & \textbf{59.31} & 66.62 & 22.34 & 70.47 & \textbf{26.71}\\
         128 & 94.49 & 58.95 & 65.03 & 24.07 & 69.73 & 25.37\\
         64 & 95.10 & 56.86 & 68.62 & \textbf{24.73} & 70.10 & 22.55 \\
         32 & 95.34 & 57.35 & 62.50 & 20.48 & 67.77 & 22.18 \\ 
         16 & 94.49 & 56.49 & 64.89 & 19.41 & 65.69 & 19.12 \\ 
         \hline
    \end{tabular}}
    \caption{Recall results for different dimensionalities of the original MinkUNeXt global descriptor.}
    \label{tab:dimensionality_results}
\end{table}

The findings presented in this study are of considerable interest for place recognition tasks in repetitive settings. They demonstrate that a more compact descriptor can achieve equivalent performance. This is significant because vineyards are unstructured and comparatively less informative than urban settings. Therefore, mapping all the salient features does not require a high-dimensional descriptor. Furthermore, selecting a high output channel value for these algorithms in agricultural settings can result in noise encoding because these settings lack of a significant number of features. For example, when choosing a lower output dimensionality of $64$ instead of $256$ increases Recall@1 increases by $+ 4\%$ when evaluated in May. This suggests that the computational load can be significantly reduced while maintaining uncompromised or even enhancing performance. However, selecting a very low output channel value for the algorithm's output (such as $32$ or $16$) begins to have a detrimental impact on the overall results and evaluation. Therefore, selecting an output dimensionality that is too low for our algorithm would be a poor choice.

Following this trend, we decided to use a final descriptor dimension of $192$, as the performance is not compromised and it reports greater Recall@1 results in two of the three evaluations proposed.

\subsection{Impact of design choices}\label{subsec:multiloss_design} 
To justify the proposed architecture, we assess the impact of different design choices concerning the use of a multi-loss approach and architectural pruning of the base method. As it was done in Section \ref{subsec:dimensionality_results}, the experimental setup used for this ablation study is training on the TEMPO-VINE's dataset February campaigns ("run1" and "run2") with the Livox sensor's data as an input. The division of the data into training and test sets was done by using the "run1" trajectory as the training set and the "run2" trajectory as the test set, as outlined in Section \ref{sec:experimental_setup}. The other hyperparameters for the experiments are set to the specified in the MinkUNeXt work \cite{cabrera2025minkunext}, except for the total number of training epochs, which is set to 200 (half of the number proposed in the cited paper) because the datasets used in this work are considerably smaller.

able \ref{tab:design_changes_results} shows the impact on the network's performance after implementing the various design decisions outlined in Section \ref{sec:methodology} is demonstrated. The specific design choices that were evaluated in this ablation study are as follows:

\begin{itemize}
    \item Baseline and pre-processing: We first establish a baseline and evaluate the impact of basic parameters, specifically focusing on voxel quantization size and the necessity of coordinates normalization (IDs 1–3).

    \item Loss function selection: Subsequently, we evaluate the impact of the training objective by comparing standard approaches (TSAP and Contrastive Loss, IDs 1-3, 4) against multi-loss setups (IDs 5-7), ultimately highlighting the benefits of adopting a MRL approach (IDs 8-9).

    \item Architectural layer selection: Finally, we investigate the impact of the descriptor extraction point. We compare the performance of extracting the descriptor from the standard final layer (Global) versus our proposed pruned architecture (Intermediate) (IDs 6–9).
\end{itemize}

\begin{figure}[h]
    \centering
    \includegraphics[width=\linewidth]{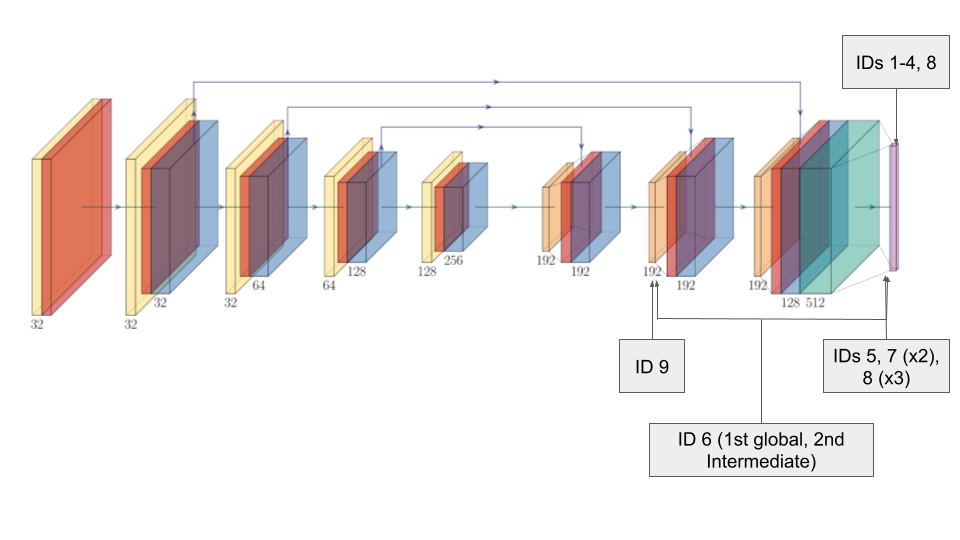}
    \caption{Design choices evaluated in the ablation study. Each number corresponds to a specific experiment identified by its ID, which is detailed in Table \ref{tab:design_changes_results}.}
    \label{fig:experiments_design}
\end{figure}

Figure \ref{fig:experiments_design} illustrates the different experiments conducted in this ablation study, highlighting the specific design choices that were modified in each case. The "x2" notation indicates that the same descriptor is used for both loss components in the multi-loss experiments. Similarly, the "x3" notation indicates that three descriptors are used in the multi-loss experiments. The "global" and "intermediate" labels refer to the layer from which the descriptor is extracted for loss computation, with "global" corresponding to the final layer of the original MinkUNeXt architecture and "intermediate" corresponding to the output of our proposed pruned architecture.

    \begin{table}[h]
        \centering
        \scriptsize
        {
        \begin{tabular}{c|c|c|c|c|c|c|c}
            \toprule 
             ID & Loss & Dims. & \makecell{Second\\descriptor} & \makecell{Quantiz.\\size} & Normalization & R@1\% & R@1 \\
             \midrule 
             1 & TSAP & 512 & No & 0.01 & No & 31.99 & 6.13\\
             2 & TSAP & 512 & No & 0.1 & No & 85.54 & 49.63\\
             3 & TSAP & 512 & No & 0.01 & Yes & 95.59 & 58.82\\
             4 & Contrastive Loss & 512 & No & 0.01 & Yes & 48.65 & 10.17\\
             5 & Multi-loss TSAP (2) & 192 & Global & 0.01 & Yes & 95.34 & 62.38\\
             6 & Multi-loss TSAP (2) & 192 & Intermediate & 0.01 & Yes & 97.06 & 62.62\\
             7 & Multi-loss TSAP (3) & 192 & Global & 0.01 & Yes & 93.75 & 54.04\\ 
             8 & Matryoshka & 192 & Global & 0.01 & Yes & 96.32 & 61.89\\
             9 & Matryoshka & 192 & Intermediate & 0.01 & Yes & \textbf{97.30} & \textbf{67.77}\\
             \bottomrule
        \end{tabular}}
        \caption{Results in terms of recall for each of the distinct design combinations that were tested using the MinkUNeXt method.}
        \label{tab:design_changes_results}
    \end{table}

As illustrated in Table \ref{tab:design_changes_results}, the most significant impact is observed when the quantization size is adapted to align with the scale of the coordinates. When quantization size is set to $0.01$, as is the case with the original MinkUNeXt \cite{cabrera2025minkunext} for both the baseline and refined protocols, the recall results are notably low. However, when adapting the quantization size to the unprocessed state of the coordinates and using a higher $0.1$ value, the recall results improve by $+40\%$. However, optimal outcomes are achieved by adapting the raw coordinates of the Livox sensor to a normalized $[-1, 1]$ space. Applying this preprocessing step to the raw coordinates of the point cloud achieves Recall@1\% values above 90 percent. This substantial improvement suggests that normalizing the input space ensures a voxelization density that is compatible with the network's receptive field, preventing both excessive sparsity and information loss due to collision. By scaling the relative distances to a standard range, the sparse convolution kernels can extract geometric features more effectively, maintaining structural integrity across the discretized grid.

In terms of loss function adaptation, the experiments show that the optimal learning is achieved by keeping the TSAP and not substituting it with a simple triplet contrastive loss function. This superiority is due to the fact that TSAP acts as a direct, differentiable surrogate for the evaluation metric itself (Average Precision), optimizing the global ranking of the retrieved candidates. While Triplet Loss focuses on local pairwise constraints—pushing negatives away from positives by a fixed margin—it relies heavily on complex hard-negative mining strategies to be effective. Conversely, TSAP considers the entire batch distribution simultaneously, enforcing a global ordering where positive matches are prioritized over negatives, thereby learning a more robust and structured embedding space without the instability associated with triplet selection. Adopting this loss function is critical in vineyard scenarios. The high degree of perceptual aliasing inherent to these semi-structured environments often causes triplet-based approaches to fail by enforcing separation between distinct locations that are visually indistinguishable, leading to poor convergence. Therefore, the loss function employed by our MinkUNeXt-VINE is the original TSAP approach.

Experiments in Table \ref{tab:design_changes_results} with IDs 5 through 7 show the results of using a weighted sum of multiple loss functions in the multi-loss variant. The experiments with IDs 5 and 6 consist of using the TSAP function twice. In the fifth experiment, the second component of the sum is the global descriptor; in the sixth, it is the output of the second transposed convolution. The ID 7 experiment, on the other hand, involves three operands: the original global descriptor, the TSAP function with the global descriptor (as in ID 5), and a third operand with the batch hard contrastive loss, which uses the final output of the MinkUNeXt backbone. Of these three cases, ID 6 yields the best results. Including triplet loss (ID 7) considerably degrades the output. However, the best multi-loss approach is introducing the MRL approach, as seen in the eighth and ninth experiments. Configurations 6 and 9 both use a multi-loss approach with earlier outputs. Using MRL represents a $+5\%$ improvement in the Recall@1 metric. Recall@1 is the most important metric in place recognition because it represents the success of retrieving the first candidate. These experiments demonstrate that the MRL multi-loss approach provides better performance in our LPR algorithm, despite being simpler and more computationally efficient because it does not require the full descriptor to be used multiple times.

From an architectural perspective, Table \ref{tab:design_changes_results} shows that using the output of the second transposed convolution as the final layer gives better results. Configurations 5 and 6 show that using these intermediate features as the second descriptor in the multi-loss approach improves Recall@1\% by $+2\%$. However, the greatest improvement is seen when comparing IDs 8 and 9. Using these intermediate outputs as the final descriptor for the MRL approach results in a significant improvement of almost $+6\%$ in the Recall@1 metric. 

\begin{figure}[h]
    \centering
    \includegraphics[width=\linewidth]{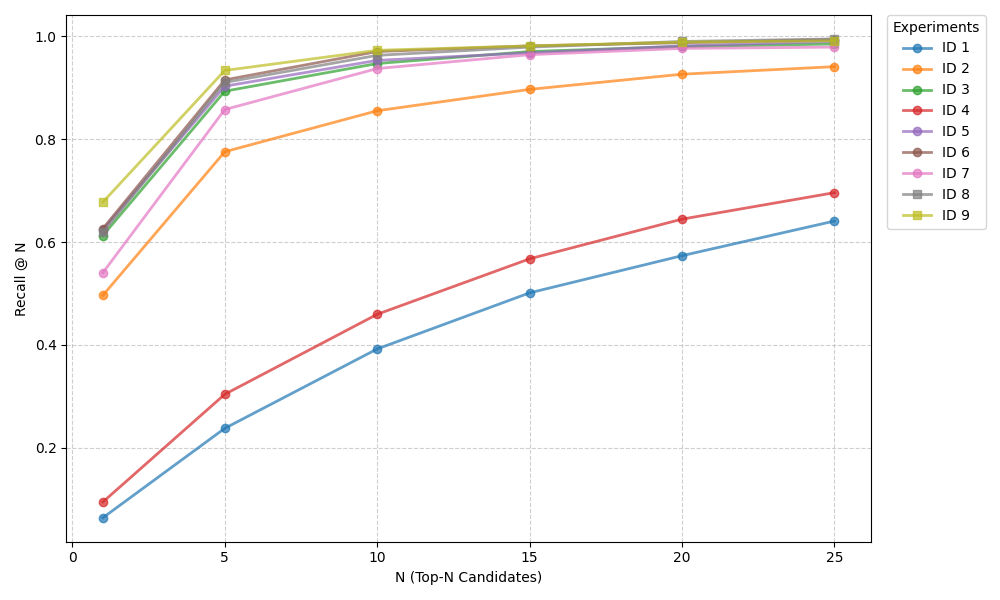}
    \caption{Recall@N graph for the different design decisions tested.}
    \label{fig:recall_design_changes}
\end{figure}

Therefore, the configuration used in the experiment with ID 9 was the final proposal for our MinkUNeXt-VINE method because it produced better results. When compared to the original backbone (ID 2), our method improved by around $+12\%$ and $+18\%$ in the Recall@1\% and Recall@1 metrics, respectively. Figure \ref{fig:recall_design_changes} displays the Recall@N graph for all of the cases displayed in Table \ref{tab:design_changes_results}.

\subsection{Seasonal changes}\label{subsec:seasons}
For the quantitative evaluation of the impact of seasonal changes in agricultural environments on LPR tasks, we used the MinkUNeXt backbone \cite{cabrera2025minkunext} and the TEMPO-VINE dataset \cite{martini2025tempo}. The distribution of the train and test examples has been made following the "one in, one out" distribution exposed in Section \ref{sec:experimental_setup}.

\begin{figure}[h]
    \centering
    \begin{subfigure}{}
        \centering
        \includegraphics[width=0.45\textwidth]{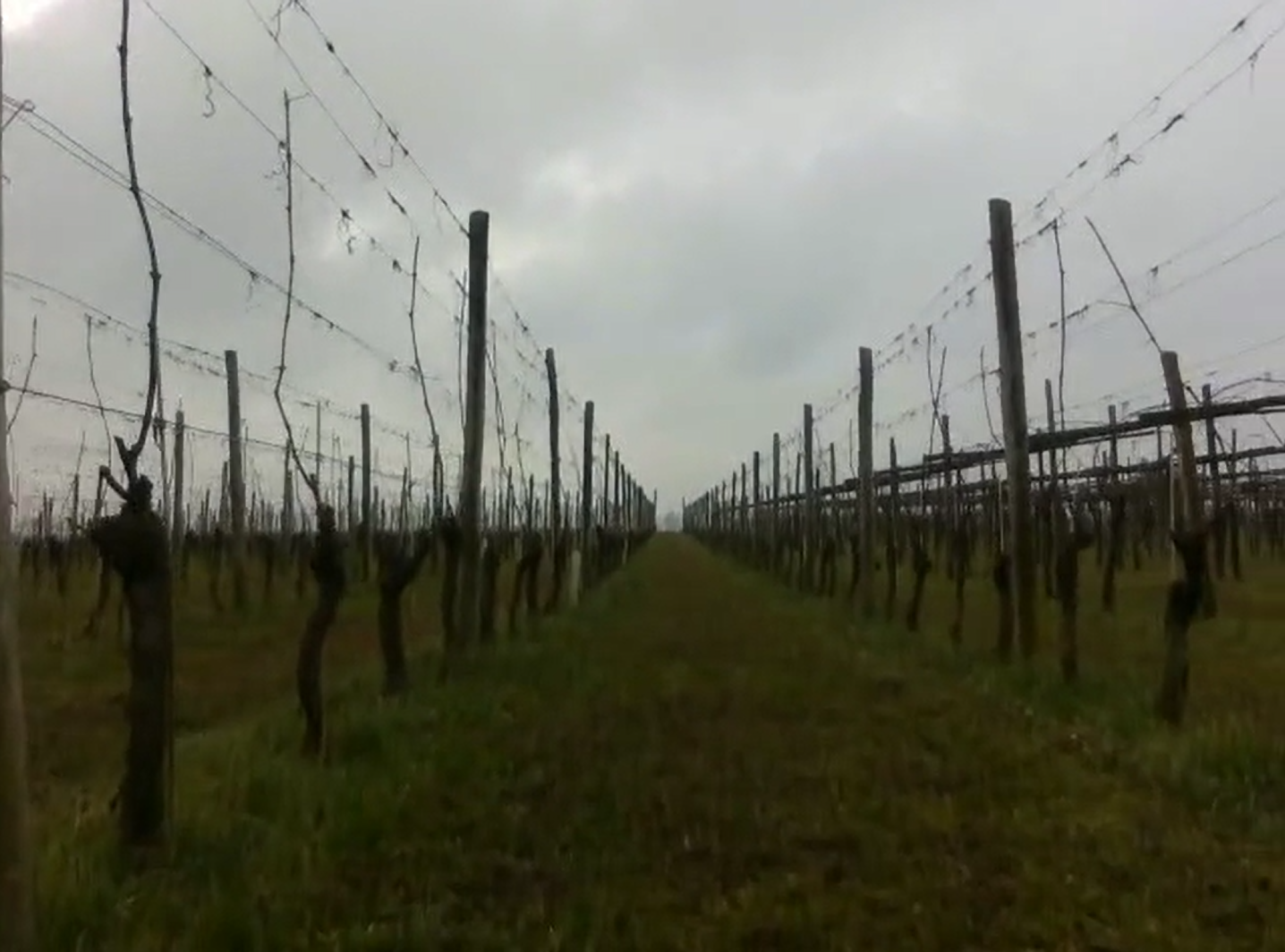}
    \end{subfigure}
    \begin{subfigure}{}
        \centering
        \includegraphics[width=0.45\textwidth]{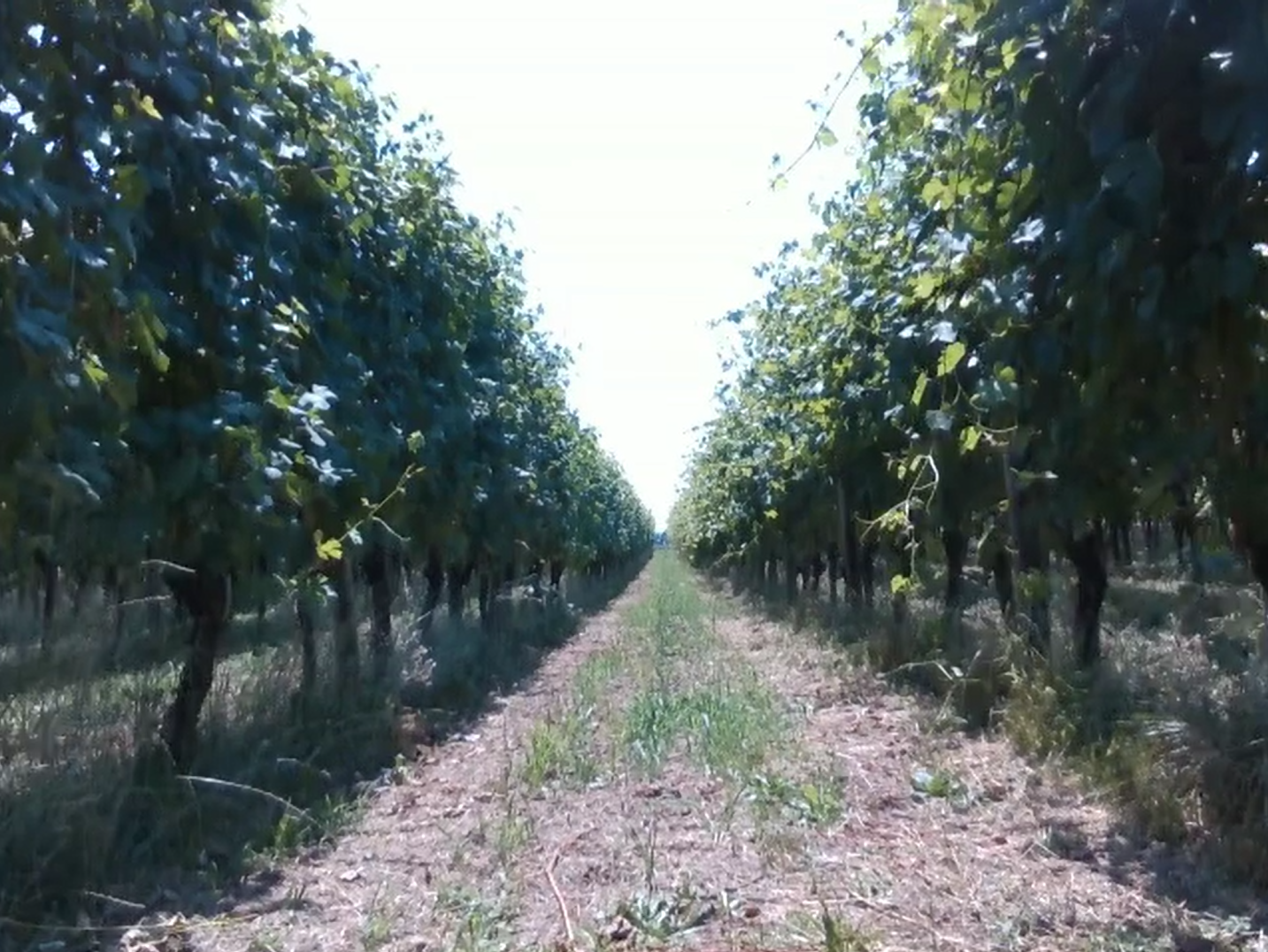}
    \end{subfigure}
    \caption{Example of seasonal changes from a same location in a vineyard environment extracted from the TEMPO-VINE dataset. Left:  Winter season with bare vines. Right:  Summer season with abundant vegetation and the grown crop.}
    \label{fig:seasonal_changes}
\end{figure}

The experiments were conducted following these different approaches:
\begin{itemize}
    \item Use of a winter sequence (February) as train set and database.
    \item Use of a spring sequence (March) as train set and database.
    \item Use of a summer sequence (the second sequence from June) as train set and database.
    \item Use of a autumn sequence (September) as train set and database.
    \item Use of season packages of three month each as train set and database.
\end{itemize}
Figure \ref{fig:seasonal_changes} represents the challenge of obtaining an accurate place recognition solution in vineyard settings. 

Tables \ref{tab:seasons_ablation_study_months_feb}, \ref{tab:seasons_ablation_study_months_march}, \ref{tab:seasons_ablation_study_months_may}, \ref{tab:seasons_ablation_study_months_june} and \ref{tab:seasons_ablation_study_months_sept} present the results obtained from implementing our MinkUNeXt-VINE proposed method for individual months: February, March, May, the end of June and September, respectively. These tables also display the generalization of the method for months that have not been trained with. Each of the selected campaigns belongs to a different season and represents a different crop stage, making them distinctly different from each other.

\begin{table}[h] 
    \centering
    \begin{tabular}{c|c|c|c} 
        \toprule
         Train set & Test set & Recall@1\% & Recall@1 \\
         \midrule 
         February & February & \textbf{85.98} & \textbf{63.26}\\ 
         February & March & 84.99 & 58.86 \\
         February & May & 58.76 & 23.12\\
         February & June (last) & 61.52 & 21.91\\
         February & September & 70.71 & 34.91\\
         \bottomrule
    \end{tabular}
    \caption{Validation results of MinkUNeXt-VINE trained with the February campaign.}
    \label{tab:seasons_ablation_study_months_feb}
\end{table}

\begin{table}[h] 
    \centering
    \begin{tabular}{c|c|c|c} 
        \toprule
         Train set & Test set & Recall@1\% & Recall@1 \\
         \midrule
         March & February & \textbf{84.91} & 56.08\\
         March & March & 83.97 & \textbf{56.44}\\
         March & May & 52.97 & 18.38\\
         March & June (last) & 61.33 & 20.80\\
         March & September & 72.28 & 33.29\\
         \bottomrule
    \end{tabular}
    \caption{Validation results of MinkUNeXt-VINE trained with the March campaign.}
    \label{tab:seasons_ablation_study_months_march}
\end{table}

\begin{table}[h] 
    \centering
    \begin{tabular}{c|c|c|c} 
        \toprule
         Train set & Test set & Recall@1\% & Recall@1 \\
         \midrule
         May & February & \textbf{78.04} & 29.89\\
         May & March & 75.11 & \textbf{30.03}\\
         May & May & 67.77 & 28.54\\
         May & June (last) & 67.32 & 24.61\\
         May & September & 64.64 & 27.03\\
         \bottomrule
    \end{tabular}
    \caption{Validation results of MinkUNeXt-VINE trained with the May campaign.}
    \label{tab:seasons_ablation_study_months_may}
\end{table}

\begin{table}[h] 
    \centering
    \begin{tabular}{c|c|c|c} 
        \toprule
         Train set & Test set & Recall@1\% & Recall@1 \\
         \midrule
         June (last) & February & 73.03 & 26.78\\
         June (last) & March & 67.62 & 27.21\\
         June (last) & May & 58.34 & 20.94\\
         June (last) & June (last) & \textbf{75.90} & \textbf{34.73}\\
         June (last) & September & 68.23 & 29.17\\
         \bottomrule
    \end{tabular}
    \caption{Validation results of MinkUNeXt-VINE trained with the end of June campaign.}
    \label{tab:seasons_ablation_study_months_june}
\end{table}

\begin{table}[h] 
    \centering
    \begin{tabular}{c|c|c|c} 
        \toprule
         Train set & Test set & Recall@1\% & Recall@1 \\
         \midrule 
         September & February & \textbf{79.46} & 41.45\\
         September & March & 78.84 & 43.61\\
         September & May & 56.99 & 19.40\\
         September & June (last) & 66.91 & 29.08\\
         September & September & 79.44 & \textbf{51.96}\\
         \bottomrule
    \end{tabular}
    \caption{Validation results of MinkUNeXt-VINE trained with the September campaign.}
    \label{tab:seasons_ablation_study_months_sept}
\end{table}
As evidenced by the five validation tables, performance remains consistent across months where the crop status is stable. However, a significant performance decay is observed in May. As reported in the TEMPO-VINE dataset article \cite{martini2025tempo}, this period is characterized by tall grass which obstructs the field of view of the Livox sensor, given its low mounting position on the Husky UGV base. Consequently, the results in Table \ref{tab:seasons_ablation_study_months_may} are lower than those of earlier months like February and March.

It is also notable that in months with fully developed canopies, such as June and September, environmental occlusions significantly increase the difficulty of the place recognition task. Comparing Table \ref{tab:seasons_ablation_study_months_feb} (February) with Table \ref{tab:seasons_ablation_study_months_june} (June), this occlusion challenge translates to a decay of approximately $-10\%$, even in the best-case scenario where the evaluation is performed on the same month used for training. In February, when the vineyard is dormant, the best results are reported, with more than 85\% of Recall@1\% and 63.26\% of Recall@1.

Furthermore, it is worth highlighting that models trained during early crop stages (February/March) generalize better to September than to June, despite the greater temporal distance. This phenomenon occurs because June represents the peak of vegetative occlusion. In contrast, September is influenced by pre-harvest interventions, specifically canopy management and leaf removal, which re-expose structural elements of the vineyard, making it geometrically more similar to the winter months.

Regarding cross-season validation, Table \ref{tab:seasons_ablation_study_seasons} presents the seasonal generalization results on three differentiated packages. Each package contains three campaigns. The "Early-Year" package contains the campaigns from February, March and April. The "Mid-Year" package contains the campaigns from May and June (both). The "Late-Year" package contains the campaigns from July (both), August and September.

\begin{table}[h] 
    \centering
    \begin{tabular}{c|c|c|c}
        \toprule
         Train set & Test set & Recall@1\% & Recall@1 \\
         \midrule 
         Early-Year & Early-Year & \textbf{78.98} & \textbf{48.07}\\
         Early-Year & Mid-Year & 49.25 & 14.82\\
         Early-Year & Late-Year & 57.75 & 24.60\\
         \midrule
         Mid-Year & Early-Year & 69.09 & 27.81\\
         Mid-Year & Mid-Year & 61.59 & 24.34\\
         Mid-Year & Late-Year & 59.44 & 26.99\\
         \midrule
         Late-Year & Early-Year & 62.07 & 28.08\\
         Late-Year & Mid-Year & 48.42 & 16.90\\
         Late-Year & Late-Year & 68.76 & 36.80\\
         \bottomrule
    \end{tabular}
    \caption{Generalization results of our proposed method on seasonal packages of TEMPO-VINE campaigns.}
    \label{tab:seasons_ablation_study_seasons}
\end{table}

From the results presented in Table \ref{tab:seasons_ablation_study_seasons}, it can be observed that, considering several campaigns into the training set can make the place recognition task harder. The reason behind this behaviour is the increased intra-class variability introduced by diverse environmental conditions. As the model is forced to map significantly different visual appearances (such as the phenological changes in the vineyard) to the same geographical location, the resulting feature representations may become less discriminative. This leads to potential confusion between distinct but visually similar places (perceptual aliasing).

Firstly, the highest performance is consistently achieved when the model is trained and tested on the same seasonal package (the diagonal of the table). Specifically, the "Early-Year" model achieves the best overall results, with a Recall@1\% of 78.98\% and a Recall@1 of 48.07\%. This suggests that the visual features during the period from February to April are more consistent or contain more distinctive landmarks that the model can easily learn, as the occlusions are not prominent in the vineyard.

Secondly, it can be noted that the Mid-Year package yields the most challenging results in both generalization from different campaigns and training. This is because the training set contains two campaigns recorded with tall grass, which introduced difficulty in the learning process of MinkUNeXt-VINE. The performance drops to Recall@1 values of 14.82\% and 16.90\% when evaluating on the Early-Year and Late-Year training scenarios, respectively. Also, the Recall@1\ metric on the Mid-Year training set does not exceed 28\% in any case.

Finally, the Late-Year training scenario presents consistent results. The generalization results of the Early-Year package presents Recall@1\% results are similar to the training case, above 60\%, but the Recall@1 metric suffers a more drastic decay of -10\% in the generalization case.

\begin{figure}[h]
    \centering
    \includegraphics[width=0.75\linewidth]{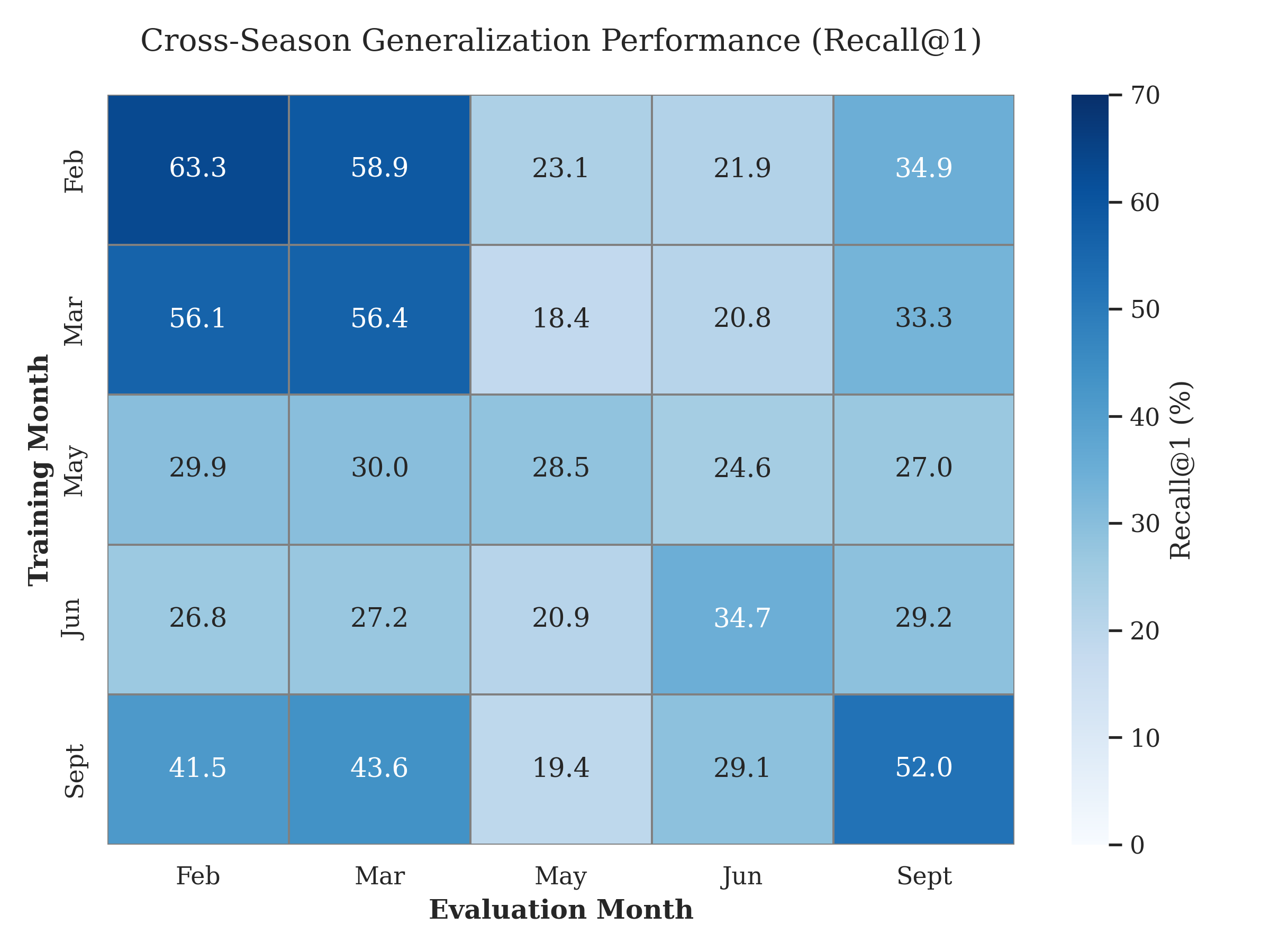}
    \caption{Cross-season generalization performance matrix showing Recall@1 values.}
    \label{fig:heatmap_seasons}
\end{figure}

Overall, the consistent gap between Recall@1\% and Recall@1 across all experiments underscores the challenge of high-precision localization; while the model can often retrieve the correct area (coarse recognition), pinpointing the exact location remains difficult under seasonal transitions. Figure \ref{fig:heatmap_seasons} presents a heatmap comparison of the results obtained in each of the individual month experiment with each generalization scenario.

\section{Results}\label{sec:results} 
\subsection{Evaluation on different architectures}\label{subsec:BLT_res}
For evaluating different place recognition approaches on long-term vineyard environments, we used the Greek campaigns of the Bacchus Long-Term (BLT) dataset \cite{polvara2024bacchus} and the campaigns from February to September from the TEMPO-VINE dataset \cite{martini2025tempo} with both of the sensors provided. The point clouds from the BLT dataset are obtained with an Ouster OS1-16 LiDAR scan. The experiments presented in this section were executed using an Nvidia GeForce RTX 3090 GPU except for the MinkUNeXt-VINE training on the TEMPO-VINE dataset, which were executed on an Nvidia A30 as made in Section \ref{sec:ablation_study}. We follow the exact same setup proposed in each of their corresponding papers for the hyperparameter selection for each of these pipelines.

Table \ref{tab:backbones} displays the results on the different backbones in terms of Recall@1\% and Recall@1. It is important to note that the hyperparameters for each of these backbones have been set according to the values specified in their corresponding papers. For these experiments, we selected a variety of methods that combine deep learning and handcrafted approaches. Primary works such as PointNetVLAD \cite{uy2018pointnetvlad} and LPD-Net \cite{liu2019lpd} are considered, as well as MinkLoc3Dv2 \cite{komorowski2022improving} for a Minkowski convolutions-based counterpart to our proposed MinkUNeXt-VINE. 

The handcrafted descriptors used in these experiments are Scan Context \cite{kim2018scan} and Fast Point Feature Histogram (FPFH) \cite{rusu2009fast}. Minimum changes were made to adapt these methods to the vineyard datasets. For the Scan Context method, the voxel size was set to 0.1 meters to account for the smaller dimensions of the vineyard datasets compared to city datasets. Following the proposed methodology, the maximum range of the points was also set to 60 meters. For FPFH, since it produces local descriptors, they were fused via global mean pooling to generate the global descriptor needed for the place recognition task. The parameters used to compute the global descriptor were a voxel size of 0.15 meters and a radius of 0.5 meters for the computed normals. Both of these handcrafted descriptors are evaluated using cosine similarity, considering a correct match to be a distance of 5 meters or less. This distance threshold is also used to evaluate every learned method in this paper, given the smaller field dimensions recorded in agricultural datasets.

\begin{landscape}
    \begin{table}[h]
        \centering
        \begin{tabular}{|c|c| >{\centering\arraybackslash}m{2cm}| >{\centering\arraybackslash}m{2cm}| >{\centering\arraybackslash}m{2cm}| >{\centering\arraybackslash}m{2cm}| >{\centering\arraybackslash}m{2cm}| >{\centering\arraybackslash}m{2cm}|}
        \hline
        \multirow{2}{*}{\textbf{Method}} & 
        \multirow{2}{*}{\textbf{Type}} & 
        \multicolumn{2}{c|}{\textbf{BLT}} & 
        \multicolumn{2}{c|}{\textbf{TEMPO-VINE (VELO)}} & 
        \multicolumn{2}{c|}{\textbf{TEMPO-VINE (Livox)}} \\ 
        \cline{3-8}
        & & Recall@1\% & Recall@1 & Recall@1\% & Recall@1 & Recall@1\% & Recall@1 \\
        \hline
            PointNetVLAD \cite{uy2018pointnetvlad} & Learned & 47.97 & - & 22.23 & - & 17.04 & -\\ 
            LPD-Net \cite{liu2019lpd} & Learned & 31.49 & - & 17.26 & - & 18.38 & -\\
            Scan Context \cite{kim2018scan} & Handcrafted & \textbf{79.84} & \textbf{40.01} & 29.64 & 7.96 & 23.05 & 4.07\\
            FPFH \cite{rusu2009fast} & Handcrafted & 72.10 & 22.87 & 27.78 & 3.95 & 25.09 & 1.41\\
            MinkLoc3Dv2 \cite{komorowski2022improving} & Learned & 36.05 & 17.73 & 14.53 & 2.14 & 12.51 & 1.61\\
            MinkUNeXt \cite{cabrera2025minkunext} & Learned & 58.12 & 30.49 & 30.49 & 11.62 & 23.08 & 6.31 \\
            MinkUNeXt-VINE (ours) & Learned & 56.32 & 25.23 & \textbf{69.71} & \textbf{41.54} & \textbf{55.57} & \textbf{25.03} \\ 
            \bottomrule
        \end{tabular}
        \caption{Evaluation of different LPR pipelines using the full extension of the different LiDAR place recognition datasets available in the literature.}
        \label{tab:backbones}
    \end{table}    
\end{landscape}

Quantitative results presented in Table \ref{tab:backbones} demonstrate the capabilities of the proposed MinkUNeXt-VINE architecture across different sensor modalities and environments.

Our method's primary contribution is evident in the TEMPO-VINE dataset. Unlike the BLT dataset, this dataset provides more challenging 3D data for place recognition because it was recorded with a higher linear velocity in the robotic platform and registers different heights of grass in the field. In this setting, MinkUNeXt-VINE significantly outperforms all competing methods. Specifically, using the Velodyne sensor, our method achieves a Recall@1\% of $69.71\%$, representing a dramatic improvement over the best-performing baseline (Scan Context at $29.64\%$). Furthermore, in the highly challenging setup using the Livox sensor, characterized by its non-repetitive scanning pattern and sparsity, our method achieves a Recall@1\% of $55.57\%$, whereas traditional descriptors like PointNetVLAD and FPFH fail to surpass $26\%$. 

The enhancement is particularly notable in comparison with its predecessor, MinkUNeXt. With our proposed architectural and preprocessing changes, our method achieves a $+39\%$ improvement in Recall@1\% and a $+32\%$ improvement in Recall@1 for the Velodyne sensor, and a $+30\%$ improvement in Recall@1\% and a $+19\%$ improvement in Recall@1 for the Livox sensor, respectively, for the same input. This confirms that our architecture successfully learns robust global descriptors resilient to the severe geometric distortions and occlusions typical of vineyard environments across the full phenological cycle.

On the BLT dataset, MinkUNeXt-VINE achieves a Recall@1\% of 56.32\% and a Recall@1 of 25.23\%, underperforming the vanilla MinkUNeXt baseline (58.12\% and 30.49\%). As suggested by the architectural ablation, the simplifications that allow our method to achieve state-of-the-art results on the sparse TEMPO-VINE datasets introduce a specific trade-off: the model sacrifices fine-grained discrimination to maintain high robustness under sparsity.

Because the BLT dataset features denser point clouds from an Ouster sensor and is characterized by distinct, static landmarks near vineyard boundaries, it heavily favors methods that exploit high-resolution structural details. This is clearly reflected in the superior performance of Scan Context, which achieves a Recall@1\% of 79.84\%. Handcrafted methods excel here by relying on distinct keypoints, but as our TEMPO-VINE results show, they fail to generalize to the ambiguous geometry of open fields. Therefore, the minor degradation on BLT is an acceptable, inherent consequence of designing a network optimized for highly repetitive, sparse agricultural environments.


\subsection{Computational analysis}\label{subsec:mem_analysis}
In this section, we analyze the inference time and the size of the deep learning models evaluated in Section \ref{subsec:BLT_res}. Handcrafted descriptors such as Scan Context \cite{kim2018scan} and FPFH \cite{rusu2009fast} are not included in this analysis because they do not require a training phase and their inference time is negligible compared to deep learning-based methods.

\begin{figure}[h]
    \centering
    \includegraphics[width=\linewidth]{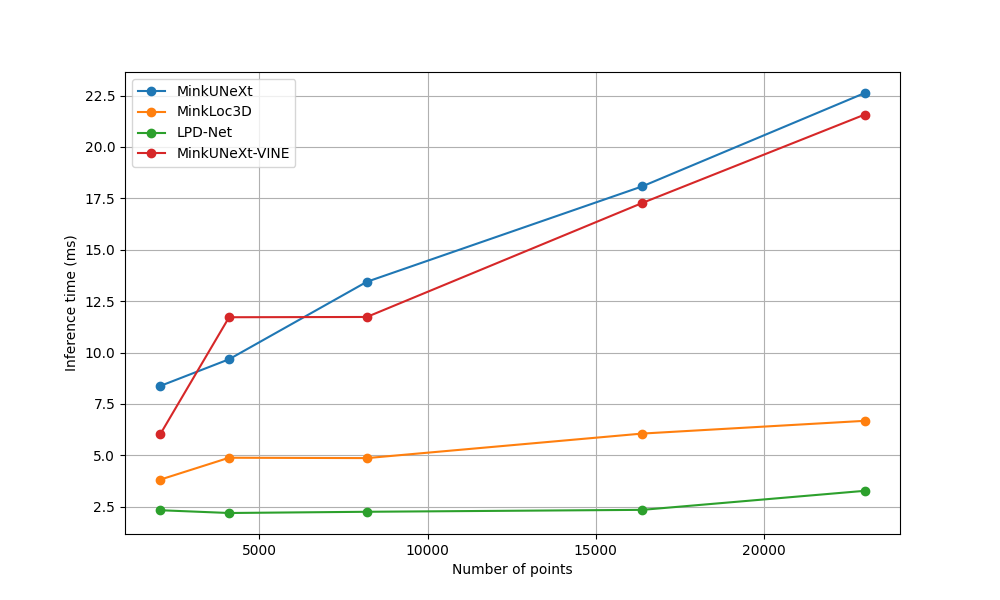}
    \caption{Inference time for a single point cloud with each of the pipelines analyzed in this study.}
    \label{fig:inference}
\end{figure}

Figure \ref{fig:inference} shows a graph with the inference time for each of the deep learning-based pipelines considered in this section at different point cloud resolutions. All of the computations were made on a GPU Nvidia RTX 3090. Note that PointNetVLAD \cite{uy2018pointnetvlad} is not included, as the inference time was calculated during evaluation and the official repository does not provide a set of pretrained weights or a code approach for saving the weights.

From Figure \ref{fig:inference}, we can conclude that, while the MinkUNeXt \cite{cabrera2025minkunext} pipeline show better performance in Table \ref{tab:backbones}, it also reports higher inference times. The number of points in the point clouds has a significant impact on this value. Finally, we conclude that structures such as LPD-Net \cite{liu2019lpd} that rely solely on the geometric properties of the point cloud and a fixed point cloud size result in the shortest inference delay. The irregular memory pattern of sparse convolutions contributes to the final inference computation. 

Regarding our proposed MinkUNeXt-VINE, the quantitative results demonstrate a distinct advantage in computational stability. As shown in Figure \ref{fig:inference}, the inference time stabilizes between 4,096 and 8,192 points, maintaining a consistent latency of approximately 11.7 ms. Furthermore, the method scales more efficiently than its predecessor (MinkUNeXt) as point density increases. Given that the architecture is optimized for sparse inputs, this stable execution time ensures suitability for real-time applications requiring predictable latency.

Furthermore, Table \ref{tab:params} presents the quantity of parameters utilized for each of the methods used in Section \ref{sec:results}. This metric serves as a crucial indicator of the computational and memory requirements associated with deploying and training these models. 

\begin{table}[h]
    \centering
    \begin{tabular}{c|c}
        \hline
         Method & Number of parameters (Millions) \\
         \hline
         PointNetVLAD \cite{uy2018pointnetvlad} & 19.8 \\
         LPD-Net \cite{liu2019lpd} & 19.8 \\
         MinkLoc3Dv2 \cite{komorowski2022improving} & 2.7 \\
         MinkUNeXt \cite{cabrera2025minkunext} & 43.5 \\
         MinkUNeXt-VINE (ours) & 27.8 \\
         \hline
    \end{tabular}
    \caption{Number of parameters in millions for each of the deep learning methods tested on vineyard environments.}
    \label{tab:params}
\end{table}

It can be observed that, while having around half the parameters of the base pipeline, MinkUNeXt, and not being considerable larger than baseline methods such as PointNetVLAD or LPD-Net, MinkUNeXt-VINE produces the best outcome in vineyard environments with sparse inputs.

\section{Conclusion}\label{sec:conclusion}
In this work, we address the problem of LiDAR place recognition in vineyard environments. The unstructured nature of these environments, characterized by the lack of distinctive features and repetitive patterns, makes finding a solution to the localization problem highly complex. Our research yields the following primary contributions to the field of agricultural robotics:
\begin{itemize}
    \item We demonstrate that the precise processing of 3D coordinates is more critical in unstructured vineyard rows than in structured urban environments.

    \item We justify the transition toward simpler deep learning architectures. These models produce lower-dimensional global descriptors that are more effective at encoding repetitive settings while reducing computational overhead.

    \item By incorporating a MRL multi-loss function, we successfully boosted the discriminativeness of descriptors. This approach proved vital for maintaining robustness when dealing with the sparse inputs typical of low-cost sensors.
\end{itemize}

To test our benchmark, we used the only two long-term vineyard datasets with LiDAR information recorded in vineyards to this day: the BLT and the TEMPO-VINE dataset. The long-term results demonstrate the resilience of the proposed architecture against sparsity and low resolution inherent to non-repetitive scanning patterns provided by low-cost sensors, such as the Livox in the TEMPO-VINE dataset. It also remains resilient in the presence of noisy input, such as data collected in an open field at higher velocities with varying grass heights, as seen in the TEMPO-VINE dataset with both Velodyne and Livox sensors.


As future work, we intend to investigate multi-modal sensor fusion strategies, exploiting the availability of camera and LiDAR data in the datasets to further improve robustness. Additionally, we target the deployment of these techniques in a real-world setting, bridging the gap between offline validation and onboard real-time operation.

\section*{Author statement}
We would like to make the following declarations: All authors confirm that this work is original and has not been published elsewhere, nor is it currently under consideration for publication elsewhere.

We confirm that the manuscript has been read and approved by all named authors and that there are no other persons who satisfy the criteria for authorship but are not listed. We further confirm that the order of authors listed in the manuscript has been approved by all of us.

We understand that the Corresponding Author is the sole contact for the Editorial process. He is responsible for communicating with the other authors about progress, submissions of revisions and final approval of proofs.

\section*{CRediT authorship contribution statement}
\textbf{Judith Vilella-Cantos:} Writing – review \& editing, Writing – original draft, Software, Methodology, Data curation, Conceptualization. \textbf{Mauro Martini:} Supervision, Formal analysis, Conceptualization. \textbf{Marcello Chiaberge:} Resources, Conceptualization, Project administration. \textbf{Mónica Ballesta:} Supervision, Formal analysis, Writing – review \& editing, Project administration, Funding acquisition. \textbf{David Valiente:} Supervision, Formal analysis, Writing – review \& editing, Project administration, Funding acquisition.

\section*{Declaration of competing interest}
The authors declare that they have no known competing financial interests or personal relationships that could have appeared to influence the work reported in this paper.

\section*{Acknowledgements}
This research work is part of the project PID2023-149575OB-I00 funded by MICIU/AEI/10.13039/501100011033 and by FEDER, UE. It is also part of the project CIPROM/2024/8, funded by Generalitat Valenciana, Conselleria de Educación, Cultura, Universidades y Empleo (program PROMETEO 2025).

\section*{Data availability}
The data of our model is available for reproduction on a public Github repository. This data can be accessed through the following link: https://git\-hub.com/JudithV/MinkUNeXt-VINE.


\end{document}